\def\year{2022}\relax
\documentclass[letterpaper]{article} 
\usepackage{aaai22}  
\usepackage{times}  
\usepackage{helvet}  
\usepackage{courier}  
\usepackage[hyphens]{url}  
\usepackage{graphicx} 
\usepackage{natbib}  
\usepackage{caption} 
\DeclareCaptionStyle{ruled}{labelfont=normalfont,labelsep=colon,strut=off} 
\usepackage{algorithm}

\usepackage{newfloat}
\usepackage{listings}
\floatstyle{ruled}
\newfloat{listing}{tb}{lst}{}
\floatname{listing}{Listing}

\usepackage{booktabs} 
\usepackage{svg}
\usepackage{algpseudocode}
\usepackage{cite}
\usepackage{amsmath,amssymb,amsfonts}
\usepackage{amsmath}

\usepackage{textcomp}
\usepackage{xcolor}
\usepackage{todonotes}

\usepackage{hyperref}
\usepackage{cleveref}

\usepackage{pifont}
\usepackage{placeins}

\usepackage{multirow}
\usepackage{acro}
\usepackage{color}

\usepackage{siunitx}

\usepackage{listings}
\usepackage{caption}
\usepackage{subcaption}

\usepackage{mwe}

\definecolor{gray}{rgb}{0.4,0.4,0.4}
\definecolor{darkblue}{rgb}{0.0,0.0,0.6}
\definecolor{cyan}{rgb}{0.0,0.6,0.6}

\definecolor{asparagus}{rgb}{0.53, 0.66, 0.42}

\newcommand{\cifar}{CIFAR-10}

\newcommand{\cifarhun}{CIFAR-100}

\newcommand{\imagenet}{ImageNet}
\newcommand{\smallimagenet}{ImageNet-32}
\newcommand{\celebahq}{CelebaHQ-4}
\newcommand{\wideresnetcif}{WideResNet28-10}

\newcommand{\autoattack}{{\it AutoAttack}}

\newcommand{\sota}{state of the art}

\newcommand{\whitebox}{Whitebox}
\newcommand{\blackbox}{Blackbox}

\newcommand{\apgdce}{APGD-CE}
\newcommand{\apgdt}{APGD-t}

\newcommand{\squaredef}{Squares}

\DeclareAcronym{knn}{
  short=k-nn,
  long=k-nearest neighbor,
}

\DeclareAcronym{nnif}{
  short=NNIF,
  long=Nearest Neighbor and Influnce Functions,
}

\DeclareAcronym{wrn}{
  short=WRN,
  long=Wide Residual Networks,
}

\DeclareAcronym{cnn}{
  short=CNN,
  long=Convolutional Neural Networks,
}

\DeclareAcronym{at}{
  short=AT,
  long=Adversarial Training,
}

\DeclareAcronym{pca}{
  short=PCA,
  long=Principal Component Analysis,
}

\DeclareAcronym{fnr}{
  short=FNR,
  long=False Negative Rate,
}

\DeclareAcronym{asr}{
  short=ASR,
  long=Adversarial Succes Rate,
}

\DeclareAcronym{asrd}{
  short=ASRD,
  long=Adversarial Success Rate under Detection,
}

\DeclareAcronym{bb}{
  short=BB,
  long=Blackbox,
}

\DeclareAcronym{wb}{
  short=WB,
  long=Whitebox,
}

\DeclareAcronym{lid}{
  short=LID,
  long=Local Intrinsic Dimensionality,
}

\DeclareAcronym{mah}{
  short=M-D,
  long=Mahalanobis Distance,
}

\DeclareAcronym{sota}{
  short=SOTA,
  long=state-of-the-art,
}

\DeclareAcronym{dft}{
  short=DFT,
  long=Discrete Fourier Transformation,
}

\DeclareAcronym{fft}{
  short=FFT,
  long=Fast Fourier Transformation,
}

\DeclareAcronym{mfs}{
  short=MFS,
  long=magnitude Fourier spectrum,
}

\DeclareAcronym{pfs}{
  short=PFS,
  long=phase Fourier spectrum,
}

\DeclareAcronym{dnn}{
  short=DNN,
  long=Deep Neural Network,
}

\DeclareAcronym{fgsm} {
  short=FGSM,
  long=Fast Gradient Method,
}

\DeclareAcronym{bim} {
  short=BIM,
  long=Basic Iterative Method,
}

\DeclareAcronym{autoattack} {
  short=AA,
  long=AutoAttack,
}

\DeclareAcronym{pgd} {
  short=PGD,
  long=Projected Gradient Descent,
}

\DeclareAcronym{df} {
  short=DF,
  long=DeepFool,
}

\DeclareAcronym{cw} {
  short=C\&W,
  long=Carlini\&Wagner,
}

\title{Is RobustBench/AutoAttack a suitable Benchmark for Adversarial Robustness?}
\title{Is RobustBench/AutoAttack a suitable Benchmark for Adversarial Robustness?}
\author {
    Peter Lorenz\textsuperscript{\rm 1,2,3},
    Dominik Straßel\textsuperscript{\rm 1,3},
    Margret Keuper\textsuperscript{\rm 4} and
    Janis Keuper\textsuperscript{\rm 1,3,5}
}
\affiliations {
    \textsuperscript{\rm 1} Fraunhofer Research Center Machine Learning, Germany\\
    \textsuperscript{\rm 2} Heidelberg University, Germany\\
    \textsuperscript{\rm 3} Fraunhofer ITWM, Kaiserslautern, Germany\\
    \textsuperscript{\rm 4} University  of  Siegen, Max Planck Institute for Informatics, Saarland Informatics Campus, Germany\\
    \textsuperscript{\rm 5} Institute for Machine Learning and Analytics (IMLA), Offenburg University,  Germany \\
    Correspondence to peter.lorenz@itwm.fhg.de
}

\usepackage{bibentry}

\begin{document}

\maketitle

\begin{abstract}
Recently, \textit{RobustBench}~\cite{Croce2020RobustBench} has become a widely recognized benchmark for the adversarial robustness of image classification networks. In its most commonly reported sub-task, \textit{RobustBench} evaluates and ranks the adversarial robustness of trained neural networks on \textit{CIFAR-10} under AutoAttack~\cite{Croce2020ReliableEO} with $l_\infty$ perturbations limited to $\epsilon=8/255$. With leading scores of the currently best-performing models of around $60\%$ of the baseline, it is fair to characterize this benchmark to be  challenging. \\
Despite its general acceptance in recent literature, we aim to foster discussion about the suitability of \textit{RobustBench} as a key indicator for robustness which could be generalized to practical applications. Our line of argumentation against this is two-fold and supported by excessive experiments presented in this paper: We argue that I) the alternation of data by AutoAttack with $l_\infty, \epsilon=8/255$ is unrealistically strong, resulting in close to perfect detection rates of adversarial samples even by simple detection algorithms while other attack methods are much harder to detect and achieve similar success rates,  II) results on low-resolution data sets like \cifar~ do not generalize well to higher resolution images as gradient-based attacks appear to become even more detectable with increasing resolutions.     

\end{abstract}

\noindent Source code: \href{https://github.com/adverML/SpectralDef_Framework}{SpectralDefense Framework} \\

\section{Introduction}

Increasing the robustness of neural network architectures against adversarial examples in general and more specifically against coordinated adversarial attacks has recently received increasing attention. In this work, we focus on the benchmarking of robustness in the context of CNN-based computer vision models.     
\subsubsection{RobustBench. }\label{rel_autoattack}In 2020, ~\cite{Croce2020RobustBench} launched a benchmark website\footnote{robustbench.github.io} with the goal of providing a standardized benchmark for adversarial robustness on image classification models. Until then, single related libraries such as FoolBox \cite{foolbox}, Cleverhans \cite{papernot2018cleverhans} and AdverTorch \cite{2019advertorch} were already available but did not include all \ac{sota}~methods in one evaluation. \\
The current rankings in \textit{RobustBench} as well as the majority of evaluations of adversarial robustness in recent literature are dominated by \textit{RobustBench's} own attack scheme \autoattack~ \cite{Croce2020ReliableEO}. \autoattack~ is an ensemble of 4 attacks: two variations of the \ac{pgd} \cite{pgd} attack with  cross-entropy loss (\apgdce) and difference of logits ratio loss (\apgdt), the targeted version of the FAB attack \cite{fabtattack}, and the blackbox \squaredef~ attack \cite{squareattack}.


\begin{figure}[t!]
    \centering
    \includegraphics[width=0.85\columnwidth]{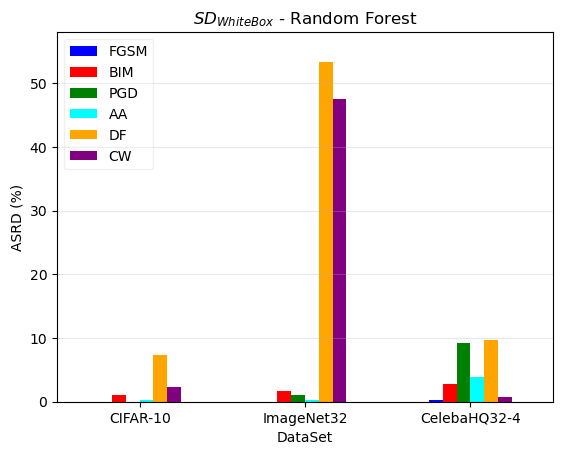}
    \caption{Attack Success Rates under Defence (ASRD) of different adversarial attack methods on several datasets for a simple defense:\ac{wb} Fourier domain detector with random forest \cite{original}: \textit{RobustBench's AutoAttack} are so easy to detect that successful attacks are very unlikely compared with other methods. \label{fig:teaser}}
\end{figure} 

\subsubsection{Contributions}
The aim of this paper is to raise the awareness that \textit{RobustBench's AutoAttack} in its default evaluation scheme $l_\infty, \epsilon=8/255$ is unrealistically strong, resulting in close to perfect detection rates of adversarial samples even by simple detection algorithms. Also, we find that benchmarks on low-resolution datasets like CIFAR-10 tend to underestimate the strength of adversarial attacks and can not be directly generalized to applications with higher resolutions. In detail, we show that:
\begin{itemize}
\item adversarial samples generated by \textit{AutoAttack} $l_\infty, \epsilon=8/255$ are modifying test images to the extent that these manipulations can easily be detected, almost entirely preventing successful attacks in practice. 
\item given a simple defense, \textit{AutoAttack} is outperformed by other existing attacks even for optimized $\epsilon$ parameters. 
\item in contrast to other methods, the effectiveness of \textit{AutoAttack} is dropping with increasing image resolutions.
\end{itemize}


\section{Methods}

\subsection{Attack Methods} \label{sec:data_generation}
For our analysis, we generate test data using \textit{AutoAttack} and a baseline of five other commonly used attack methods from the \textit{foolbox} \cite{foolbox}.  We employ the untargeted version of all attacks, if available.


\subsubsection{\acf{autoattack}:}        \textit{RobustBench} is based on the evaluation of \ac{autoattack}~ \cite{Croce2020ReliableEO}, which is an ensemble of 4 parameter-free attacks: two variations of the \ac{pgd} attack \cite{pgd} (see \Cref{sssec:pgd}) with cross-entropy loss (\apgdce) and difference of logits ratio loss (\apgdt):
        \begin{equation*}
            \text{DLR}(x,y) = \frac{z_y - \max_{x\neq y} z_i}{z_{\pi 1} - z_{\pi 3} }.
        \end{equation*}
        where $\pi$ is ordering of the components of $z$ in decreasing order. The \apgdt~ can handle models with a minimum of 4 classes.
        The targeted version of the FAB attack \cite{fabtattack}, and the \ac{bb} \squaredef~ attack \cite{squareattack}. 
        The \ac{autoattack}~framework provides two modes. \textit{RobustBench} uses the ``standard'' mode, executing the 4 attack methods consecutively. The failed attacked samples are handed over to the next attack method, to ensure a higher attack rate.

\subsubsection{\acf{fgsm}:} The \ac{fgsm} \cite{fgsm} uses the gradients of the \ac{dnn} to create adversarial examples. For an input image, the method uses the gradients of the loss w.r.t. the input image to create a new image that maximizes the loss. This output is called the adversarial image. The following expression summarizes this:
        \begin{equation*}
            X^{adv} = X -  \epsilon \text{sign}( \nabla_{X} J(X_{N}^{adv},y_{t}))\; \text{,}
        \end{equation*}
     where $X^{adv}$ is the adversarial image, $X$ is the original input image, $y$ is the original input label, $\epsilon$ is the multiplier to ensure the perturbations are small and $J$ is the loss. There is no guarantee that the generated adversarial examples by this method are similar to their real counterpart. 

\subsubsection{\acf{bim}:}
        
        The method \ac{bim} \cite{bim} is the iterative version of \ac{fgsm}. After each iteration, the pixel values need to be clipped to ensure the generated adversarial examples is still within the range of both the $\epsilon$ ball (i.e. $[x-\epsilon, x+\epsilon]$) and the input space (i.e. $[0, 255]$ for the pixel values). The formulation is expressed as follows:
        \begin{equation*}
            \begin{aligned}
                X_{0}^{adv} &= X, \\
                X_{N+1}^{adv} &= \text{CLIP}_{X,\epsilon} \{ X_{N}^{adv} - \alpha \text{sign}( \nabla_{X} J(X,y_{t})) \},
            \end{aligned}
        \end{equation*}
        where $N$ denotes the number of iterations. 

\subsubsection{\acf{pgd}:\label{sssec:pgd}}
        The \ac{pgd} \cite{pgd}  is a variant of \ac{bim} and one of the most popular whitebox (allowing full  access to model gradients and weights) attacks. It introduces random initialization of the perturbations for each iteration. This algorithm strives to find the perturbation that maximizes a model's loss on a particular input. The size of the perturbation is kept smaller than an amount by $\epsilon$. This constraint is expressed ether as $l_2$ or $l_\infty$ norm. 

\subsubsection{\acf{df}:}
        The \ac{df} is a non-targed method that is able to find the minimal amount of perturbations possible which mislead the model using an iterative linearization approach \cite{deepfool}. The main idea is to find the closest distance from the input sample to the model decision boundary. 

\subsubsection{\ac{cw}:}

        The attack method \acf{cw} \cite{cw}  is based on the L-BFGS and has three versions: $l_0$, $l_2$, and $l_\infty$. We employ the $l^2$ variant which is most commonly used. This attack method generates for an given input $X$ an adversarial example $X^{adv}$ by formulating following optimization problem:
        \begin{equation*}
        \begin{aligned}
             \min  \lVert \frac{1}{2} (\tanh(X^{adv}) + 1) - X \rVert + c f(\frac{1}{2} (\tanh(X^{adv}) + 1)) \\
             \text{With }  f(x) = \max(Z(x)_{true} - \max_{i \neq true} \{Z(x)_i \},0),
        \end{aligned}
        \end{equation*}
        
    where $Z(x)$ is the softmax classification result vector. The initial value for $c$ is $c=10^{-3}$ , a binary search is performed to then find the smallest $c$, s.t.  $f(X_{adv}) \leq 0$. 

\subsection{Measuring the Success of Adversarial Attacks }
{\it RobustBench}, like most of the benchmarks in the literature regarding adversarial robustness, uses a \textit{Robust Accuracy}~\cite{Croce2020RobustBench}~ measure to compare different methods. However, this approach does not fit our evaluation scheme, since we are aiming to measure the success of adversarial samples under defense to obtain a more realistic view of the practical impact of the applied attacks. Therefore, we reformulate the robustness measures and report two different indicators: 

\paragraph{Attack Success Rate (ASR)}

    The {\it \ac{asr}} in \cref{eq:asr}  is calculated as
    \begin{equation}
        \text{ASR} = \frac{ \text{\#~perturbed~samples }}{ \text{\#~all~samples} } \label{eq:asr}
    \end{equation}
     the fraction of successfully perturbed test images and it provides a baseline of an attacker's ability to fool unprotected target networks. Hence, {\it \ac{asr}} is providing the same information as \textit{Robust Accuracy} from an attacker's perspective.

\paragraph{Attack Success Rate under Defense (ASRD)}

We extend {\it \ac{asr}} by the practical assumption that too strong perturbations can be detected at inference time. To measure the performance of attacks under defense, we introduce the {\it \ac{asrd} } in \cref{eq:asrd}, computing the ratio of successful attacks
\begin{equation}
    \text{ASRD} = \frac{ \text{\#~undetected~perturbations} } { \text{\#~all~samples} } = \text{FNR} \cdot \text{ASR,} \label{eq:asrd}
\end{equation}
where \Acs{fnr} is the false negative rate of the applied detection algorithm.

\subsection{A Simple Adversarial Detector}
In order to measure the magnitude of perturbations imposed by \textit{RobustBench}, we apply a simple and easy-to-implement adversarial detector, called SpectralDefense (SD), introduced in \cite{original, lorenz2021detecting}. 
This method is based on a feature extraction in the Fourier domain, followed by a \textit{Logistic Regression} or \textit{Random Forest} classifier. 
It can be applied in a blackbox fashion, using only the (adversarial) input images, or as whitebox detector accessing the feature maps of attacked neural networks. 
In both cases, the detector is based on a Fourier transformation \cite{fft}:   
	For a discrete 2D signal, like color image channels or single CNN feature maps -- $X\in[0,1]^{N\times N}$ -- the 2D discrete Fourier transform is given as
	\begin{equation}\label{eq:eq1}
	    \mathcal{F}(X)(l,k) = \sum_{n,m=0}^N e^{-2\pi i \frac{lm+kn}{N}}X(m,n),
	\end{equation}
	for $l,k = 0,\ldots N-1$, with complex valued Fourier coefficients $\mathcal{F}(X)(l,k)$.
	The detector then only utilizes the magnitudes of Fourier coefficients 
	\begin{equation}
	    |\mathcal{F}(X)(l,k)| = \sqrt{\text{Re}(\mathcal{F}(X)(l,k))^2 +\text{Im}(\mathcal{F}(X)(l,k))^2}
	    \label{eq:fftabs}
	\end{equation}
	to detect adversarial attacks with high accuracy.

\subsubsection{\blackbox~Detection: Fourier Features of Input Images}
    While different attacks show distinct but randomly located change patterns in the spatial domain (which makes them hard to detect), \cite{original} showed that adversarial samples have strong, well-localized signals in the frequency domain. \\
	Hence, the detector extracts and concatenates the 2D power spectrum of each color channel as feature representations of input images and uses simple classifiers like \textit{Random Forests} and \textit{Logistic Regression} to learn to detect perturbed input images.
	
	\subsubsection{\whitebox~ Detection: Fourier Features of Feature-Maps}
	In the whitebox case, the detector applies the same method as in the blackbox approach but extends the inputs to the feature map responses of the target network to test samples. Since this extension will drastically increase the feature space for larger target networks, only a subset of the available feature maps are selected. 
    In original paper \cite{original} and in the follow-up paper \cite{lorenz2021detecting}, it is stated that a combination of several layers delivers better detection results.

\section{Experiments} \label{sec:exp}
    
    Since most of the successful methods ranked on \textit{Robustbench} are based on a \wideresnetcif~\cite{wideresidual} architecture, we also conduct our evaluation on a baseline \wideresnetcif~ using the following datasets without applying adversarial examples or other methods to increase the robustness during training. \\

    \subsubsection{\cifar.}
    We train on the plain \cifar~training set to a test accuracy of 87\% and apply the different attacks on the test set. Then, we extract the spectral features and use a random subset of 1500 samples of this data for each attack method to evaluate {\it \ac{asr} } and {\it \ac{asrd} }. 
    
    \subsubsection{\cifarhun. }
    The procedure is similar to \cifar~ dataset. We train on the \cifarhun~training set to a test accuracy of 79\% and apply the attacks on the test set. 
    
    \subsubsection{\smallimagenet. (64 and 128.)}
    This dataset~\cite{imagenet32} (and its variants $64\times 64$ and $128\times 128$ pixels) has the exact same number of classes (1000) and images as the original \imagenet~with the only difference that the images are downsampled. Moreover, a lower resolution of the images makes the classification task more difficult and the baseline test accuracy is 66\%  and 77\% respectively. 
    
    \subsubsection{\celebahq-32. (64 and 128.)}
    This dataset~\cite{celebahq} provides images of celebrities' faces in HQ quality ($1024\times 1024px$) whereas we downsampled it to $32$, $64$ and $128$ pixels width and height. We only selected the attributes ``Brown Hair'', ``Blonde Hair'', ``Black Hair'' and ``Gray Hair'' to train the \ac{wrn} to a test accuracy of 91\%. The data is unbalanced, where the class ``Gray Hair'' has the least samples.  

\subsection{Detecting Attacks}
\begin{figure}[h!]
    \centering
    \includegraphics[width=0.85\columnwidth]{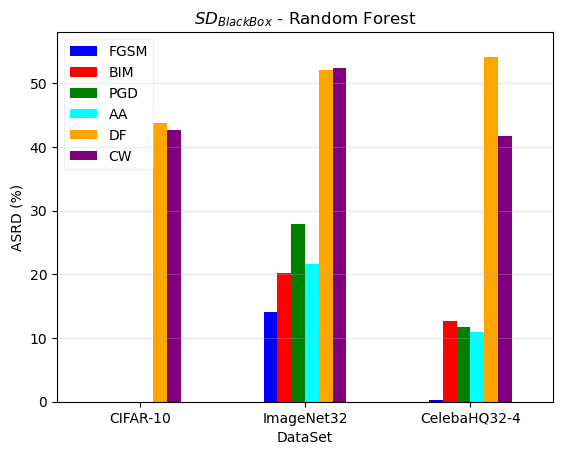}
    \caption{Blackbox ASRD comparison using a Random Forest classifier on different $32\times 32$ datasets.\label{fig:ASRD-32}}
\end{figure}

Figures \ref{fig:teaser} and \ref{fig:ASRD-32} show a subset of whitebox and blackbox ASRD results for all attack methods on datasets with a resolution of  $32\times 32$\footnote{The full ASRD evaluation on all datasets is listed in table \cref{tab:appendixallnets} of the appendix.}. In both cases, \textit{AutoAttack} has very low ASRD rates, not only compared to other methods but also in absolute values. In most cases, the probability of successful \ac{autoattack} attacks is marginally low.   

\subsection{AutoAttack for different choices of $\epsilon$}
\begin{figure}[H]
    \centering
    \includegraphics[width=0.85\columnwidth]{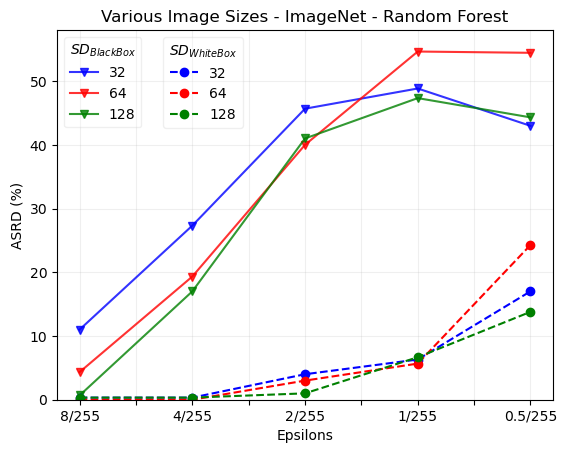}
    \caption{ASRD of AA with random forest for a range of different $\epsilon$ on ImageNet.\label{fig:eps}}
\end{figure}

\begin{figure}[H]
    \centering
    \includegraphics[width=0.85\columnwidth]{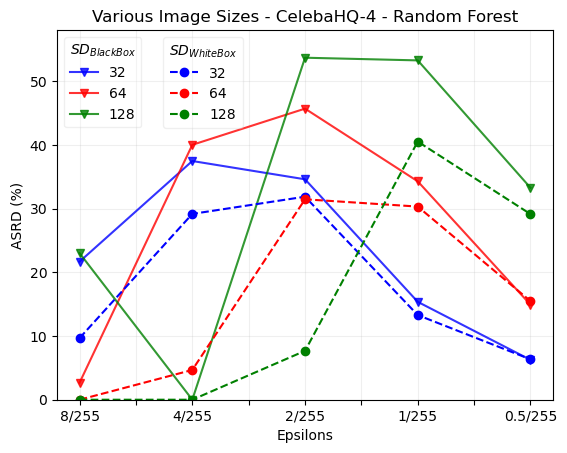}
    \caption{ASRD of AA with random forest for a range of different $\epsilon$ on CelebHQ.\label{fig:eps2}}
\end{figure}

One might argue that the low \ac{asrd} rates of \ac{autoattack} might be caused by too high choice of $\epsilon$. Hence, we repeat the full set of \textit{AutoAttack} experiments for a full range of different $\epsilon$-values. Figures \ref{fig:eps} and \ref{fig:eps2} show a subset of this evaluation for ImageNet and CelebHQ on different $\epsilon$, image resolutions as well as WB and BB detectors with Random Forests\footnote{Full evaluation results in table \Cref{tab:appendixallepsilons} of the appendix. }.

\FloatBarrier
\subsection{Success Rates depending on Image Resolution}
\begin{figure}[H]
    \centering 
    \includegraphics[width=0.85\columnwidth]{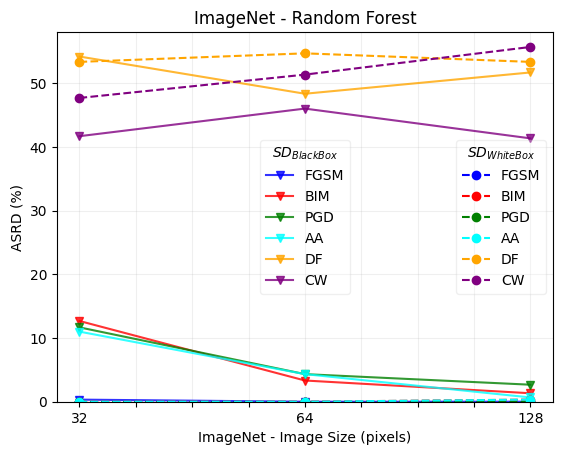}
    \caption{ ASRD with Random Forest classifiers on increasing resolutions of \imagenet.}
    \label{fig:in_resolution}
\end{figure}

\begin{figure}[H]
    \centering
    \includegraphics[width=0.85\columnwidth]{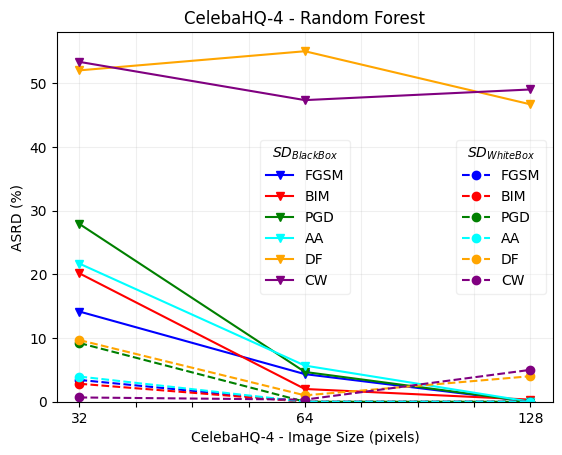}
    \caption{ ASRD with Random Forest classifiers on increasing resolutions of \celebahq-4.}
    \label{fig:celeba_resolution}
\end{figure}

As shown in \Cref{fig:in_resolution} and \ref{fig:celeba_resolution}, we compare the \ac{asrd} over the three image size ($s = \{32, 64, 128\}$) on the datasets \celebahq~ and \imagenet.  The attacks \ac{fgsm}, \ac{bim}, \ac{pgd}, and \ac{autoattack} are sensitive to the image size. The used detector has better results as the image size is increased. In contrast, \ac{df} and \ac{cw} keep their attack strength overall image sizes $s$. Again, \ac{autoattack} does not show sufficient results for using adversarial detection robustness.  

 \FloatBarrier
\section{Discussion}
The results of our empirical evaluations show strong evidence that the widely used \autoattack~ scheme for benchmarking the adversarial robustness of image classifier models on low-resolution data might not be a suitable setup to generalize the obtained results to estimate the robustness in practical vision applications. Even for lower choices of the $\epsilon$-parameter,    \autoattack~ still appears to modify target images beyond reasonable class boundaries. Additionally, the resolution of the benchmark images should not be neglected. In terms of resolution as well as in the number of classes and training images,  \cifar~ is a conveniently sized dataset for the very expensive \sota~ adversarial training approaches. However, our experiments suggest that these results might not generalize to more complex problems.\\  
In light of our results, we argue that too strong adversarial benchmarks like the current setting of \textit{RobustBench} might hamper the development of otherwise practically relevant methods towards more model robustness.

\FloatBarrier

\bibstyle{aaai22}

\captionsetup[table]{name=\textbf{Appendix}}

\begin{table*}
\centering
\resizebox{\linewidth}{!}{%
\begin{tabular}{|c|l|r|rrrrrr|rrrrrr|} 
\hline
\multicolumn{2}{|c|}{\multirow{3}{*}{\textbf{Arch: Wide ResNet 28-10}}} & \multicolumn{1}{c|}{\multirow{3}{*}{\textbf{ASR}}} & \multicolumn{6}{c|}{\textbf{SD$_\text{BlackBox}$}} & \multicolumn{6}{c|}{\textbf{SD$_\text{WhiteBox}$}} \\ 
\cline{4-15}
\multicolumn{2}{|c|}{} & \multicolumn{1}{c|}{} & \multicolumn{2}{c|}{F1} & \multicolumn{2}{c|}{FNR} & \multicolumn{2}{c|}{ASRD} & \multicolumn{2}{c|}{F1} & \multicolumn{2}{c|}{FNR} & \multicolumn{2}{c|}{ASRD} \\ 
\cline{4-15}
\multicolumn{2}{|c|}{} & \multicolumn{1}{c|}{} & \multicolumn{1}{c|}{LR} & \multicolumn{1}{c|}{RF} & \multicolumn{1}{c|}{LR} & \multicolumn{1}{c|}{RF} & \multicolumn{1}{c|}{LR} & \multicolumn{1}{c|}{RF} & \multicolumn{1}{c|}{LR} & \multicolumn{1}{c|}{RF} & \multicolumn{1}{c|}{LR} & \multicolumn{1}{c|}{RF} & \multicolumn{1}{c|}{LR} & \multicolumn{1}{c|}{RF} \\ 
\hline
\multirow{6}{*}{\textbf{CIFAR-10}} & FGSM & 95.08 & 97.34 & 97.72 & 2.33 & 0.00 & 2.22 & 0.00 & 99.01 & 97.88 & 0.00 & 0.00 & 0.00 & 0.00 \\ 
\cline{2-2}
 & BIM & 99.37 & 92.93 & 95.54 & 8.00 & 0.00 & 7.95 & 0.00 & 97.65 & 96.44 & 3.00 & 0.67 & 2.98 & 0.67 \\ \cline{2-2}
 & PGD & 99.27 & 91.79 & 95.24 & 8.67 & 0.00 & 8.61 & 0.00 & 96.70 & 95.85 & 2.33 & 0.00 & 2.31 & 0.00 \\ \cline{2-2}
 & AA  & 100.0 & 91.78 & 96.31 & 7.00 & 0.00 & 7.00 & 0.00 & 98.00 & 96.76 & 2.00 & 0.33 & 2.00 & 0.33 \\ \cline{2-2}
 & DF & 100.0 & 48.31 & 49.47 & 54.67 & 53.33 & 54.67 & 53.33 & 54.42 & 52.30 & 45.67 & 47.00 & 45.67 & 47.00 \\ \cline{2-2}
 & CW & 100.0 & 48.07 & 53.75 & 54.33 & 42.67 & 54.33 & 42.67 & 53.29 & 54.52 & 47.33 & 40.67 & 47.33 & 40.67 \\ \hline
\multirow{6}{*}{\textbf{CIFAR-100}} 
& FGSM & 99.95 & 94.58 & 97.72 & 7.00 & 0.00 & 7.00 & 0.00 & 99.34 & 98.85 & 0.33 & 0.00 & 0.33 & 0.00 \\ \cline{2-2}
 & BIM & 99.95 & 87.39 & 95.39 & 15.67 & 0.00 & 15.66 & 0.00 & 97.00 & 98.50 & 3.00 & 1.33 & 3.00 & 1.33 \\ \cline{2-2}
 & PGD & 99.95 & 86.97 & 95.24 & 14.33 & 0.00 & 14.32 & 0.00 & 96.83 & 98.68 & 3.33 & 0.00 & 3.33 & 0.00 \\ \cline{2-2}
 & AA  & 100.0 & 92.57 & 96.76 & 8.67 & 0.33 & 8.67 & 0.33 & 97.35 & 97.72 & 2.00 & 0.00 & 2.00 & 0.00 \\ \cline{2-2}
 & DeepFool & 100.0 & 50.17 & 51.84 & 49.67 & 46.00 & 49.67 & 46.00 & 50.33 & 48.00 & 49.33 & 54.00 & 49.33 & 54.00 \\  \cline{2-2}
 & CW & 100.0 & 50.17 & 64.20 & 49.67 & 10.33 & 49.67 & 10.33 & 47.92 & 47.29 & 54.00 & 55.00 & 54.00 & 55.00 \\ \hline
\multirow{6}{*}{\textbf{ImageNet32}} 
& FGSM        & 99.95 & 84.53 & 90.20 & 15.33 & 0.33 & 15.32 & 0.33 & 100.0 & 99.83 & 0.00 & 0.00 & 0.00 & 0.00 \\ \cline{2-2}
 & BIM        & 100.0 & 71.33 & 78.68 & 30.33 & 12.67 & 30.33 & 12.67 & 100.0 & 99.67 & 0.00 & 0.33 & 0.00 & 0.33 \\ \cline{2-2}
 & PGD        & 100.0 & 74.70 & 78.75 & 26.67 & 11.67 & 26.67 & 11.67 & 100.0 & 99.67 & 0.00 & 0.67 & 0.00 & 0.67 \\  \cline{2-2}
 & AA         & 100.0 & 71.74 & 79.82 & 29.33 & 11.00 & 29.33 & 11.00 & 99.67 & 99.67 & 0.00 & 0.33 & 0.00 & 0.33 \\  \cline{2-2}
 & DeepFool   & 100.0 & 66.59 & 48.45 & 0.33 & 53.00  & *  & 53.00 & 50.33 & 48.98 & 49.33 & 52.00 & 49.33 & 52.00 \\ \cline{2-2}
 & CW         & 100.0 & 66.59 & 50.82 & 0.33 & 48.33  & *  & 48.33 & 51.46 & 49.41 & 47.00 & 51.33 & 47.00 & 51.33 \\ \hline
\multirow{6}{*}{\textbf{ImageNet64}} 
& FGSM & 100.0 & 88.15 & 92.59 & 12.00 & 0.00 & 12.00 & 0.00 & 99.83 & 99.67 & 0.00 & 0.00 & 0.00 & 0.00 \\ \cline{2-2}
 & BIM & 100.0 & 74.29 & 84.30 & 26.33 & 3.33 & 26.33 & 3.33 & 99.50 & 99.17 & 0.33 & 0.00 & 0.33 & 0.00 \\ \cline{2-2}
 & PGD & 100.0 & 75.63 & 82.59 & 25.00 & 4.33 & 25.00 & 4.33 & 99.67 & 99.67 & 0.33 & 0.00 & 0.33 & 0.00 \\ \cline{2-2}
 & AA  & 100.0 & 78.54 & 81.42 & 21.33 & 4.33 & 21.33 & 4.33 & 99.83 & 99.67 & 0.00 & 0.00 & 0.00 & 0.00 \\ \cline{2-2}
 & DeepFool & 100.0 & 49.32 & 50.82 & 51.33 & 48.33 & 51.33 & 48.33 & 50.66 & 48.63 & 48.67 & 52.67 & 48.67 & 52.67 \\ \cline{2-2}
 & CW & 100.0 & 60.84 & 51.92 & 22.33 & 46.00 & * & 46.00 & 49.24 & 45.29 & 51.67 & 58.33 & 51.67 & 58.33 \\ \hline
\multirow{6}{*}{\textbf{ImageNet128}} 
& FGSM        & 100.0 & 89.55 & 92.88 & 10.00 &  0.00 & 10.00 &  0.00 & 99.83 & 99.34 &  0.00 &  0.00 &  0.00 &  0.00 \\ \cline{2-2}
 & BIM        & 100.0 & 81.43 & 91.36 & 20.33 &  1.33 & 20.33 &  1.33 & 99.50 & 98.52 &  0.00 &  0.33 &  0.00 &  0.33 \\ \cline{2-2}
 & PGD        & 100.0 & 81.82 & 90.82 & 19.00 &  2.67 & 19.00 &  2.67 & 99.67 & 99.34 &  0.00 &  0.00 &  0.00 &  0.00 \\ \cline{2-2}
 & AA         & 100.0 & 77.34 & 85.51 & 18.67 &  0.67 & 18.67 &  0.67 & 99.34 & 98.19 &  0.00 &  0.33 &  0.00 &  0.33 \\ \cline{2-2}
 & DeepFool   & 100.0 & 66.67 & 49.15 &  0.00 & 51.67 &  * & 51.67 & 53.85 & 51.61 & 41.67 & 46.67 & 41.67 & 46.67 \\ \cline{2-2}
 & CW         & 100.0 & 60.00 & 53.99 & 25.00 & 41.33 &  * & 41.33 & 54.41 & 48.19 & 40.33 & 53.33 & 40.33 & 53.33 \\ \hline
\multirow{6}{*}{\textbf{CelebaHQ32-4}}
 & FGSM & 78.59 & 75.95 & 76.64 & 23.67 & 18.00 & 18.60 & 14.15 & 85.95 & 93.44 & 13.33 & 5.00 & 10.48 & 3.93 \\ \cline{2-2}
 & BIM & 95.91 & 73.97 & 74.06 & 22.33 & 21.00 & 21.42 & 20.14 & 84.48 & 96.35 & 12.00 & 3.33 & 11.51 & 3.19 \\ \cline{2-2}
 & PGD & 90.93 & 71.40 & 68.99 & 29.67 & 30.67 & 26.98 & 27.89 & 79.47 & 91.46 & 20.00 & 9.00 & 18.19 & 8.18 \\ \cline{2-2}
 & AA  & 100.0 & 69.49 & 74.25 & 31.67 & 21.67 & 31.67 & 21.67 & 87.79 & 88.71 & 11.33 & 9.67 & 11.33 & 9.67 \\ \cline{2-2}
 & DeepFool & 100.0 & 59.05 & 49.32 & 39.67 & 52.00 & 39.67 & 52.00 & 63.59 & 57.69 & 35.67 & 49.33 & 35.67 & 49.33 \\ \cline{2-2}
 & CW & 100.0 & 55.76 & 48.64 & 44.33 & 52.33 & 44.33 & 52.33 & 61.11 & 58.46 & 37.67 & 40.67 & 37.67 & 40.67 \\ \hline
\multirow{6}{*}{\textbf{CelebaHQ64-4}} 
& FGSM & 100.0 & 93.27 & 90.97 & 5.33 & 4.33 & 5.33 & 4.33 & 98.01 & 99.67 & 1.33 & 0.33 & 1.33 & 0.33 \\ \cline{2-2}
 & BIM & 100.0 & 95.16 & 95.30 & 5.00 & 2.00 & 5.00 & 2.00 & 98.66 & 99.50 & 1.67 & 0.67 & 1.67 & 0.67 \\ \cline{2-2}
 & PGD & 100.0 & 90.85 & 91.67 & 9.00 & 4.67 & 9.00 & 4.67 & 97.17 & 99.50 & 2.67 & 0.33 & 2.67 & 0.33 \\ \cline{2-2}
 & AA  & 100.0 & 84.26 & 84.60 & 14.33 & 5.67 & 14.33 & 5.67 & 97.17 & 100.0 & 2.67 & 0.00 & 2.67 & 0.00 \\ \cline{2-2}
 & DeepFool & 100.0 & 48.08 & 47.04 & 54.00 & 55.00 & 54.00 & 55.00 & 49.31 & 49.66 & 52.33 & 51.33 & 52.33 & 51.33 \\ \cline{2-2}
 & CW & 100.0 & 50.25 & 50.89 & 50.00 & 47.33 & 50.00 & 47.33 & 50.25 & 45.58 & 50.67 & 57.00 & 50.67 & 57.00 \\ \hline
\multirow{6}{*}{\textbf{CelebaHQ128-4}}
 & FGSM    & 95.74 & 98.82 & 97.40 & 2.00 & 0.00 & 1.91 & 0.00 & 99.67 & 100.0 & 0.67 & 0.00 & 0.64 & 0.00 \\ \cline{2-2}
 & BIM     & 99.95 & 98.16 & 98.03 & 2.00 & 0.33 & 2.00 & 0.33 & 99.16 & 100.0 & 1.33 & 0.00 & 1.33 & 0.00 \\ \cline{2-2}
 & PGD      & 99.76 & 97.37 & 98.20 & 1.33 & 0.00 & 1.33 & 0.00 & 99.16 & 100.0 & 1.33 & 0.00 & 1.33 & 0.00 \\ \cline{2-2}
 & AA       & 100.0 & 93.57 & 92.88 & 3.00 & 0.00 & 3.00 & 0.00 & 98.67 & 100.0 & 1.33 & 0.00 & 1.33 & 0.00 \\ \cline{2-2}
 & DeepFool & 100.0 & 55.21 & 52.98 & 44.33 & 46.67 & 44.33 & 46.67 & 55.65 & 50.87 & 45.00 & 56.33 & 45.00 & 56.33 \\ \cline{2-2}
 & CW & 100.0 & 51.63 & 50.50 & 47.33 & 49.00 & 47.33 & 49.00 & 52.87 & 50.26 & 46.33 & 51.00 & 46.33 & 51.00 \\ \hline
\end{tabular}
}
\caption{Results of the proposed detectors on AutoAttack
(standard mode) for different choices of the hyper-parameter
$\epsilon$ (default in most publications is $\epsilon=8/255$) and test sets.
ASR=Attack Success Rate, ASRD=Attack Success Rate under Detection. \acf{bb} and \acf{wb} results on all datasets are obtained by a Logistic Regression classifier and Random Forests. F1 and the \acf{fnr} are used to report the detection performance. See \Cref{sec:exp} for details of the experimental setup. Note that \ac{asrd} values marked by a star '*' are missing values.}
\label{tab:appendixallnets}
\end{table*}

\begin{table*}
\centering
\resizebox{\linewidth}{!}{%
\begin{tabular}{|c|l|r|rrrrrr|rrrrrr|} 
\hline
\multicolumn{2}{|c|}{\multirow{3}{*}{\textbf{Arch: Wide ResNet 28-10}}} & \multicolumn{1}{c|}{\multirow{3}{*}{\textbf{ASR}}} & \multicolumn{6}{c|}{\textbf{SD$_\text{BlackBox}$}} & \multicolumn{6}{c|}{\textbf{SD$_\text{WhiteBox}$}} \\ 
\cline{4-15}
\multicolumn{2}{|c|}{} & \multicolumn{1}{c|}{} & \multicolumn{2}{c|}{F1} & \multicolumn{2}{c|}{FNR} & \multicolumn{2}{c|}{ASRD} & \multicolumn{2}{c|}{F1} & \multicolumn{2}{c|}{FNR} & \multicolumn{2}{c|}{ASRD} \\ 
\cline{4-15}
\multicolumn{2}{|c|}{} & \multicolumn{1}{c|}{} & \multicolumn{1}{c|}{LR} & \multicolumn{1}{c|}{RF} & \multicolumn{1}{c|}{LR} & \multicolumn{1}{c|}{RF} & \multicolumn{1}{c|}{LR} & \multicolumn{1}{c|}{RF} & \multicolumn{1}{c|}{LR} & \multicolumn{1}{c|}{RF} & \multicolumn{1}{c|}{LR} & \multicolumn{1}{c|}{RF} & \multicolumn{1}{c|}{LR} & \multicolumn{1}{c|}{RF} \\ 
\hline
\multirow{5}{*}{\textbf{CIAR-10}}
 & AA (8/255)   & 100.0 & 91.78 & 96.31 &  7.00 &  0.00 &  7.00 &  0.00 & 98.00 & 96.76 &  2.00 &  0.33 &  2.00 &  0.33 \\
 & AA (4/255)   & 100.0 & 83.36 & 92.28 & 15.67 &  0.33 & 15.67 &  0.33 & 91.00 & 88.75 &  7.33 &  2.67 &  7.33 &  2.67 \\
 & AA (2/255)   & 94.41 & 69.26 & 82.39 & 31.67 & 10.33 & 29.90 &  9.75 & 83.63 & 79.00 & 14.00 & 16.00 & 13.22 & 15.11 \\
 & AA (1/255)   & 56.39 & 57.93 & 69.61 & 44.00 & 26.33 & 24.81 & 14.85 & 69.32 & 62.79 & 30.33 & 33.33 & 17.10 & 18.79 \\
 & AA (0.5/255) & 23.14 & 52.67 & 41.33 & 55.52 & 10.95 & 47.33 &  9.56 & 58.55 & 50.00 & 40.67 & 51.00 &  9.41 & 11.80 \\ 
\hline
\multirow{5}{*}{\textbf{CIFAR-100}} 
 & AA (8/255)   & 100.0 & 92.57 & 96.76 & 8.67  &  0.33 &  8.67 & 0.33 & 97.35 & 97.72 & 2.00 & 0.00 & 2.00 & 0.00 \\
 & AA (4/255)   & 99.90 & 83.93 & 91.93 & 17.33 &  1.33 & 17.31 & 1.33 & 91.61 & 92.11 & 9.00 & 4.67 & 8.99 & 4.67 \\
 & AA (2/255)   & 97.28 & 72.03 & 82.30 & 31.33 &  9.33 & 30.48 & 9.08 & 83.22 & 83.81 & 15.67 & 12.00 & 15.24 & 11.67 \\
 & AA (1/255)   & 73.65 & 62.81 & 70.77 & 36.67 & 23.33 & 27.01 & 17.18 & 73.89 & 74.04 & 25.00 & 19.67 & 18.41 & 14.49 \\
 & AA (0.5/255) & 38.97 & 51.23 & 60.44 & 51.33 & 36.33 & 20.00 & 14.16 & 61.59 & 60.87 & 39.33 & 37.00 & 15.33 & 14.42 \\ 
\hline
\multirow{5}{*}{\textbf{ImageNet32}} 
 & AA (8/255)   & 100.0 & 71.74 & 79.82 & 29.33 & 11.00 & 29.33 & 11.00 & 99.67 & 99.67 &  0.00 &  0.33 &  0.00 &  0.33 \\
 & AA (4/255)   & 99.95 & 62.38 & 65.27 & 37.00 & 27.33 & 36.98 & 27.32 & 99.00 & 97.71 &  0.67 &  0.33 &  0.67 &  0.33 \\
 & AA (2/255)   & 100.0 & 56.58 & 55.54 & 42.67 & 45.67 & 42.67 & 45.67 & 96.82 & 94.27 &  3.67 &  4.00 &  3.67 &  4.00 \\
 & AA (1/255)   & 99.67 & 51.82 & 50.33 & 47.67 & 49.00 & 47.51 & 48.84 & 87.67 & 89.21 & 12.33 &  6.33 & 12.29 &  6.31 \\
 & AA (0.5/255) & 92.78 & 52.55 & 51.60 & 45.00 & 46.33 & 41.75 & 42.98 & 79.47 & 76.56 & 20.00 & 18.33 & 18.56 & 17.01 \\ 
\hline
\multirow{5}{*}{\textbf{ImageNet64}} 
 & AA (8/255)   & 100.0 & 78.54 & 81.42 & 21.33 &  4.33 & 21.33 &  4.33 & 99.83 & 99.67 &  0.00 &  0.00 &  0.00 &  0.00 \\
 & AA (4/255)   & 100.0 & 65.37 & 72.56 & 33.00 & 19.33 & 33.00 & 19.33 & 99.00 & 99.01 &  1.33 &  0.00 &  1.33 &  0.00 \\
 & AA (2/255)   & 100.0 & 58.84 & 58.06 & 39.00 & 40.00 & 39.00 & 40.00 & 97.03 & 94.02 &  2.00 &  3.00 &  2.00 &  3.00 \\
 & AA (1/255)   & 99.95 & 50.53 & 47.47 & 52.00 & 54.67 & 51.97 & 54.64 & 88.36 & 89.70 & 12.67 &  5.67 & 12.66 &  5.67 \\
 & AA (0.5/255) & 98.40 & 48.06 & 46.37 & 54.67 & 55.33 & 53.80 & 54.44 & 67.38 & 71.97 & 37.00 & 24.67 & 36.41 & 24.28 \\ 
\hline
\multirow{5}{*}{\textbf{ImageNet128}} 
 & AA (8/255)   & 100.0 & 77.34 & 85.51 & 18.67 & 18.67 & 18.67 &  0.67 & 99.34 & 98.19 &  0.00 &  0.33 &  0.00 &  0.33 \\
 & AA (4/255)   & 100.0 & 59.97 & 72.38 & 42.33 & 42.33 & 42.33 & 17.00 & 97.52 & 96.61 &  1.67 &  0.33 &  1.67 &  0.33 \\
 & AA (2/255)   & 98.47 & 54.93 & 57.28 & 44.33 & 44.33 & 44.33 & 41.00 & 92.28 & 90.00 &  6.33 &  1.00 &  6.33 &  1.00 \\
 & AA (1/255)   & 100.0 & 48.17 & 51.97 & 54.00 & 54.00 & 54.00 & 47.33 & 82.66 & 80.58 & 15.00 &  6.67 & 15.00 &  6.67 \\
 & AA (0.5/255) & 100.0 & 48.54 & 52.46 & 53.00 & 53.00 & 52.19 & 44.31 & 70.53 & 71.17 & 25.00 & 14.00 & 24.62 & 13.79 \\ 
\hline
\multirow{5}{*}{\textbf{CelebaHQ32-4}} 
 & AA (8/255)   & 100.0 & 69.49 & 74.25 & 31.67 & 21.67 & 31.67 & 21.67 & 87.79 & 88.71 & 11.33 &  9.67 & 11.33 &  9.67 \\
 & AA (4/255)   & 99.43 & 56.20 & 58.90 & 43.33 & 37.67 & 43.08 & 37.46 & 72.07 & 71.14 & 27.33 & 29.33 & 27.17 & 29.16 \\
 & AA (2/255)   & 68.26 & 51.86 & 50.43 & 49.00 & 50.67 & 33.45 & 34.59 & 59.31 & 56.24 & 40.00 & 46.67 & 27.30 & 31.86 \\
 & AA (1/255)   & 27.70 & 45.34 & 46.29 & 57.82 & 55.44 & 16.02 & 15.36 & 49.82 & 51.26 & 52.38 & 47.96 & 14.51 & 13.28 \\
 & AA (0.5/255) & 10.91 & 54.69 & 45.45 & 40.17 & 57.26 &  4.38 &  6.25 & 53.44 & 44.75 & 43.59 & 58.12 &  4.76 &  6.34 \\ 
\hline
\multirow{5}{*}{\textbf{CelebaHQ64-4}} 
 & AA (8/255)   & 100.0 & 84.26 & 86.90 & 14.33 &  2.67 & 14.33 &  2.67 & 97.17 & 100.0 &  2.67 &  0.00 &  2.67 &  0.00 \\
 & AA (4/255)   & 100.0 & 64.23 & 58.35 & 35.67 & 40.00 & 35.67 & 40.00 & 90.88 & 94.86 & 10.33 &  4.67 & 10.33 &  4.67 \\
 & AA (2/255)   & 99.31 & 55.19 & 52.60 & 43.33 & 46.00 & 43.03 & 45.68 & 72.51 & 73.61 & 28.33 & 31.67 & 28.13 & 31.45 \\
 & AA (1/255)   & 69.94 & 48.59 & 51.09 & 54.00 & 49.00 & 37.77 & 34.27 & 55.30 & 57.63 & 47.00 & 43.33 & 32.87 & 30.31 \\
 & AA (0.5/255) & 28.14 & 48.36 & 48.45 & 53.33 & 53.00 & 15.01 & 14.91 & 52.68 & 48.04 & 46.00 & 55.00 & 12.94 & 15.48 \\ 
\hline
\multirow{5}{*}{\textbf{CelebaHQ128-4}}
 & AA (8/255)   & 100.0 & 71.52 & 72.76 & 24.67 & 23.00 & 24.67 & 23.00 & 94.21 & 99.17 &  5.00 &  0.00 &  5.00 &  0.00 \\
 & AA (4/255)   & 100.0 & 93.57 & 92.88 &  3.00 &  0.00 &  3.00 &  0.00 & 98.67 & 100.0 &  1.33 &  0.00 &  1.33 &  0.00 \\
 & AA (2/255)   & 100.0 & 54.94 & 48.26 & 45.33 & 53.67 & 45.33 & 53.67 & 82.99 & 89.07 & 18.67 &  7.67 & 18.67 &  7.67 \\
 & AA (1/255)   & 98.02 & 51.51 & 47.08 & 48.67 & 54.33 & 47.71 & 53.25 & 63.18 & 60.17 & 37.67 & 41.33 & 36.92 & 40.51 \\
 & AA (0.5/255) & 61.98 & 50.74 & 48.52 & 48.67 & 53.67 & 30.17 & 33.26 & 53.22 & 53.36 & 47.67 & 47.00 & 29.55 & 29.13 \\
\hline
\end{tabular}
}
\caption{Different datasets are attacked by \autoattack~ but with a different epsilons for the perturbation. The \ac{asr} falls for different datasets.}
\label{tab:appendixallepsilons}
\end{table*}








\end{document}